\documentclass[conference]{IEEEtran}
\IEEEoverridecommandlockouts
\usepackage{cite}
\usepackage{amsmath,amssymb,amsfonts}
\usepackage{algorithmic}
\usepackage{graphicx}
\usepackage{textcomp}
\usepackage{xcolor}
\usepackage[caption=false, font=footnotesize]{subfig}
\usepackage{tikz}
\usepackage{placeins}
\usepackage{orcidlink}

\def\BibTeX{{\rm B\kern-.05em{\sc i\kern-.025em b}\kern-.08em
    T\kern-.1667em\lower.7ex\hbox{E}\kern-.125emX}}
    \makeatletter
\newcommand{\linebreakand}{%
  \end{@IEEEauthorhalign}
  \hfill\mbox{}\par
  \mbox{}\hfill\begin{@IEEEauthorhalign}
}
\makeatother
\begin{document}

\title{Robust Sliding Mode and Admittance Control of Underactuated Aerial Manipulators for Contact-Based Inspection}

% \author{\IEEEauthorblockN{Tareq Aziz Alqutami\IEEEauthorrefmark{1}, Yvan Petillot\IEEEauthorrefmark{2}, Matthew W. Dunnigan\IEEEauthorrefmark{3} and Mustafa Suphi Erden\IEEEauthorrefmark{4}}
% \IEEEauthorblockA{\IEEEauthorrefmark{1}\IEEEauthorrefmark{2}\IEEEauthorrefmark{3}\IEEEauthorrefmark{4} School of Engineering and Physical Sciences, Heriot-Watt University\\
% \IEEEauthorrefmark{1} Group Technology \& Commercialization (GT\&C), Petroliam Nasional Berhad (PETRONAS)\\
% Email: \IEEEauthorrefmark{1}ta2024@hw.ac.uk, \IEEEauthorrefmark{2}Y.R.Petillot@hw.ac.uk, \IEEEauthorrefmark{3}M.W.Dunnigan@hw.ac.uk, \IEEEauthorrefmark{4}m.s.erden@hw.ac.uk}
% }

\author{\IEEEauthorblockN{Tareq Aziz Alqutami\IEEEauthorrefmark{4} \orcidlink{0000-0001-7111-2737}}
\IEEEauthorblockA{
\textit{Heriot-Watt University}, Edinburgh, UK \\
\IEEEauthorrefmark{4}\textit{Petroliam Nasional Berhad (PETRONAS)}, Malaysia \\
ta2024@hw.ac.uk}
\and
\IEEEauthorblockN{Yvan Petillot \orcidlink{0000-0002-1596-289X}}
\IEEEauthorblockA{\textit{School of Engineering and Physical Sciences}  \\
\textit{Heriot-Watt University}\\
Edinburgh, UK \\
Y.R.Petillot@hw.ac.uk}

\linebreakand

\IEEEauthorblockN{Matthew W. Dunnigan}
\IEEEauthorblockA{\textit{School of Engineering and Physical Sciences}  \\
\textit{Heriot-Watt University}\\
Edinburgh, UK \\
M.W.Dunnigan@hw.ac.uk}
\and
\IEEEauthorblockN{Mustafa Suphi Erden \orcidlink{0000-0001-6199-9151}}
\IEEEauthorblockA{\textit{School of Engineering and Physical Sciences}  \\
\textit{Heriot-Watt University}\\
Edinburgh, UK \\
m.s.erden@hw.ac.uk}

}

%=============================================
% For IEEE copyright notice
% \IEEEoverridecommandlockouts
% \IEEEpubid{
% \begin{tikzpicture}[remember picture, overlay]
% \node[anchor=south, yshift=0.8cm] at (current page.south) {
% \parbox{\textwidth}{\small © 2026 IEEE. Personal use of this material is permitted. Permission from IEEE must be obtained for all other uses, in any current or future media, including reprinting/republishing this material for advertising or promotional purposes, creating new collective works, for resale or redistribution to servers or lists, or reuse of any copyrighted component of this work in other works.}
% };
% \end{tikzpicture}
% }
%========================================

\maketitle

\begin{abstract}
Contact-based industrial inspection requires aerial platforms to maintain stable interaction while rejecting disturbances. Underactuated aerial manipulators present control challenges due to the dynamic coupling between vehicle attitude and force generation. This paper proposes a robust control framework for an underactuated hexarotor equipped with a 1-DoF manipulator to perform sustained contact inspection. The architecture integrates integral-augmented Sliding Mode Control (SMC) for trajectory tracking with an admittance control law for force regulation. The contact force is mapped to a feedforward attitude term, while the 1-DoF arm actively compensates for the tilt to maintain surface alignment. Software-in-the-loop simulations demonstrate that the SMC-based approach achieves superior tracking and coupling rejection compared to traditional PID. Furthermore, the interaction strategy achieved precise force regulation with an RMSE of 0.12 N and was able to stably exert up to 20~N force, confirming the system’s efficacy for stable, reliable contact-based inspection.
\end{abstract}

\begin{IEEEkeywords}
Aerial Manipulator, Contact Inspection, NDT, Sliding Mode Control, Underactuated UAVs, Admittance Control.
\end{IEEEkeywords}

\section{Introduction}
Unmanned aerial vehicles (UAVs) have evolved from passive inspection tools into active platforms for Aerial Physical Interaction and Manipulation (APIM) \cite{olleroPresentFutureAerial2022}, enabling tasks such as pick-and-place \cite{olleroPresentFutureAerial2022}, pushing/pulling \cite{leeAerialManipulatorPushing2021}, writing \cite{nishioDesignControlMotion2024}, and contact-based inspection \cite{bodieOmnidirectionalTiltRotorFlying2023}.

Industrial contact tasks, such as non-destructive testing (NDT), minimize human hazards and operational downtime. However, maintaining stable surface contact remains challenging with underactuated multirotors, where force generation is inherently coupled with vehicle attitude.

To address this limitation, two main approaches have been pursued in the literature. The first involves fully-actuated UAV platforms, which employ tilted \cite{yiBacksteppingbasedSuperTwistingSliding2022,alqutami2024modelingICOM} or actively tilting rotors \cite{bodieOmnidirectionalTiltRotorFlying2023} to decouple translational and rotational dynamics. While this enables improved force regulation and interaction control, it comes at the cost of increased mechanical complexity, reduced efficiency, and higher control burden. The second approach retains underactuated UAVs and augments them with lightweight manipulators to facilitate interaction \cite{chenAdaptiveStiffnessVisual2024,meng2018_emergency_switch}. While more practical and energy-efficient, it introduces strong dynamic coupling between the vehicle and the manipulator, significantly complicating force regulation during contact.

From a control perspective, a variety of methods have been explored for aerial interaction. Classical PID control has been used for contact tasks such as emergency button pressing with underactuated UAVs \cite{meng2018_emergency_switch}, and has been further combined with admittance \cite{alqutami2024modelingICEM} and impedance \cite{bodieOmnidirectionalTiltRotorFlying2023} strategies for interaction control. Adaptive backstepping control has been applied to underactuated UAVs for manipulation tasks such as pick-and-place \cite{rafiqueAdaptiveBacksteppingControl2025}, while prescribed performance control (PPC) has been proposed to enforce transient and steady-state error bounds \cite{liangActivePhysicalInteraction2023}, and extended with disturbance observers for improved trajectory tracking \cite{liangLowComplexityPrescribedPerformance2022}. PPC has also been integrated with image-based visual servoing and variable stiffness control to achieve stable contact using underactuated platforms \cite{chenAdaptiveStiffnessVisual2024}.

Despite their advantages, these approaches present notable limitations. PID-based methods lack robustness to disturbances and model uncertainties. Backstepping-based controllers often result in complex recursive designs, which complicates real-time implementation, and rely heavily on accurate system models. PPC, while theoretically appealing, imposes rigid error constraints that can lead to control singularities and increased sensitivity to sensor noise, actuator saturation, and abrupt disturbances, potentially causing violations of the prescribed performance bounds.

Sliding Mode Control (SMC) offers a robust alternative due to its inherent ability to reject matched uncertainties and external disturbances with relatively low computational complexity \cite{villaAdaptiveSlidingMode2023}. SMC \cite{arizagaObserverbasedAdaptiveControl2024} and its variants, such as integral SMC \cite{yaoModelingSlidingMode2018}, have been investigated for trajectory tracking of fully-actuated aerial manipulators. For underactuated systems, terminal SMC (TSMC) \cite{zhaoBasedIterativeTerminal2023} and nonsingular fast terminal SMC (NFTSMC) \cite{xuAdaptiveNonsingularFast2023}, often combined with disturbance observers, have demonstrated improved tracking performance in free-flight scenarios. However, despite these advances, the application of SMC to sustained-contact tasks has yet to be explored for underactuated aerial manipulators.

This paper presents a robust control framework for continuous contact inspection using an underactuated hexarotor with a 1-DoF manipulator. The approach combines SMC for motion stabilization, a geometric force feedforward term, and admittance force control. The framework is validated in high-fidelity software-in-the-loop (SITL) simulations.

The remainder of this paper is organized as follows: Section II describes system modeling; Sections III and IV detail motion and interaction control; Section V presents SITL results; and Section VI concludes the paper.

\section{Aerial Platform Description}
\subsection{Dynamic modeling}
\begin{figure}[htbp]
\centering
\includegraphics[width=0.89\linewidth]{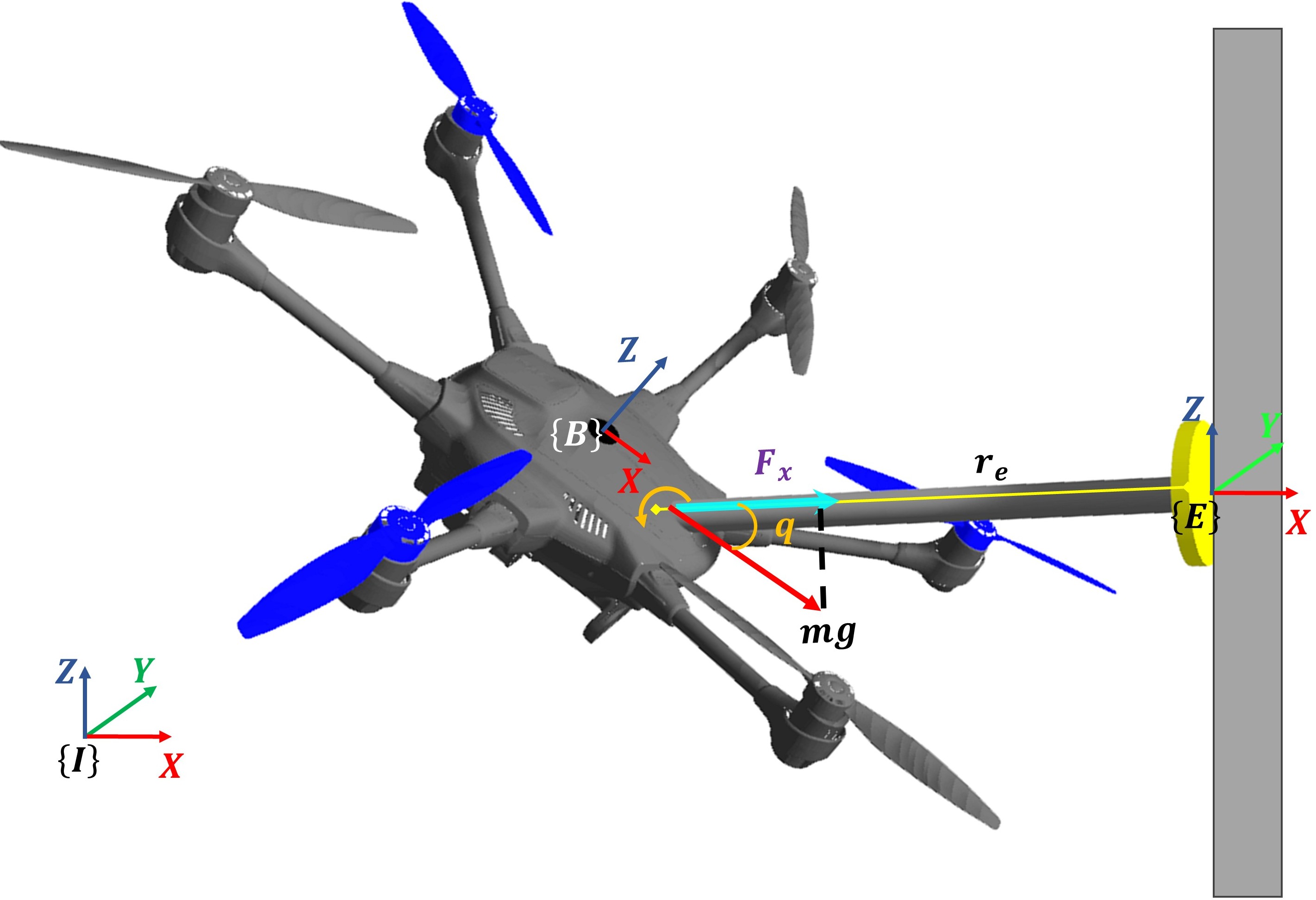}
\caption{Schematic view of the proposed aerial manipulator}
\label{fig:Schematic}
\end{figure}

The aerial manipulator comprises an underactuated hexarotor UAV and an independently actuated 1-DoF manipulator arm carrying an NDT end-effector and force sensor (Fig. \ref{fig:Schematic}). The arm actively compensates for UAV tilt to preserve sensor-surface alignment during contact.

Let $\mathcal{F}_I$, $\mathcal{F}_B$, and $\mathcal{F}_E$ denote the inertial, body-fixed, and end-effector frames, respectively. The vehicle position is $\mathbf{p} = [x, y, z]^T \in \mathcal{F}_I$, and orientation is represented by rotation matrix $\mathbf{R} \in SO(3)$ corresponding to Z-Y-X Euler angles $\Phi=[\phi,\theta,\psi]^T$, where $\phi$ is roll, $\theta$ is pitch, and $\psi$ is yaw angle. The velocity and acceleration are denoted by $\dot{\mathbf{p}}$ and $\ddot{\mathbf{p}}$ respectively, while the angular velocity in $\mathcal{F}_B$ is $\boldsymbol{\omega} = [p, q, r]^T$. The Newton-Euler equations of motion are:

\begin{IEEEeqnarray}{ll}
m \ddot{\mathbf{p}} &= m \mathbf{g} + \mathbf{R} \mathbf{f}_c + \mathbf{f}_e + \mathbf{\Delta}_f \label{eq:newton_eom}\\
\mathbf{J} \dot{\boldsymbol{\omega}} &= -\boldsymbol{\omega} \times (\mathbf{J} \boldsymbol{\omega}) + \boldsymbol{\tau}_c + \boldsymbol{\tau}_e + \boldsymbol{\Delta}_\tau \label{eq:euler_eom} \\
\dot{\mathbf{R}} &= \mathbf{R} \, \mathbf{S}(\boldsymbol{\omega}) \label{eq:rotational_eom}
\end{IEEEeqnarray}
where $m$ is mass, $\mathbf{g} = [0, 0, -g]^T$ is the gravity vector, $\mathbf{J}$ is inertia, $\mathbf{f}_c = [0, 0, T]^T$ is total thrust vector ($T=\sum_{i=1}^{n}{T_i\in \mathbb{R}}$), $\boldsymbol{\tau}_c$ is control torque, and $\mathbf{S}(\cdot)$ is the skew-symmetric matrix. More details of actuation dynamics and rotor kinematics are given in \cite{alqutami2024modelingICOM}.
$\mathbf{f}_e$ and $\boldsymbol{\tau}_e$ are external contact forces/torques. Terms $\mathbf{\Delta}_f$ and $\boldsymbol{\Delta}_\tau$ represent lumped, bounded disturbances ($\| \mathbf{\Delta}_f \| \le \sigma_f, \| \boldsymbol{\Delta}_\tau \| \le \sigma_\tau$) including arm reaction forces, aerodynamic effects, and model uncertainties. This allows treating the arm motion as a bounded disturbance in the vehicle control loop. This assumption is reasonable for lightweight mechanisms whose inertia is small relative to the vehicle mass and whose motion ($q$, and $\dot{q}$) is independently controlled. During contact inspection, the dominant interaction force acts in the normal direction of the surface. 

\subsection{System architecture}
\begin{figure*}[htbp]
\centering
\includegraphics[width=0.8\textwidth]{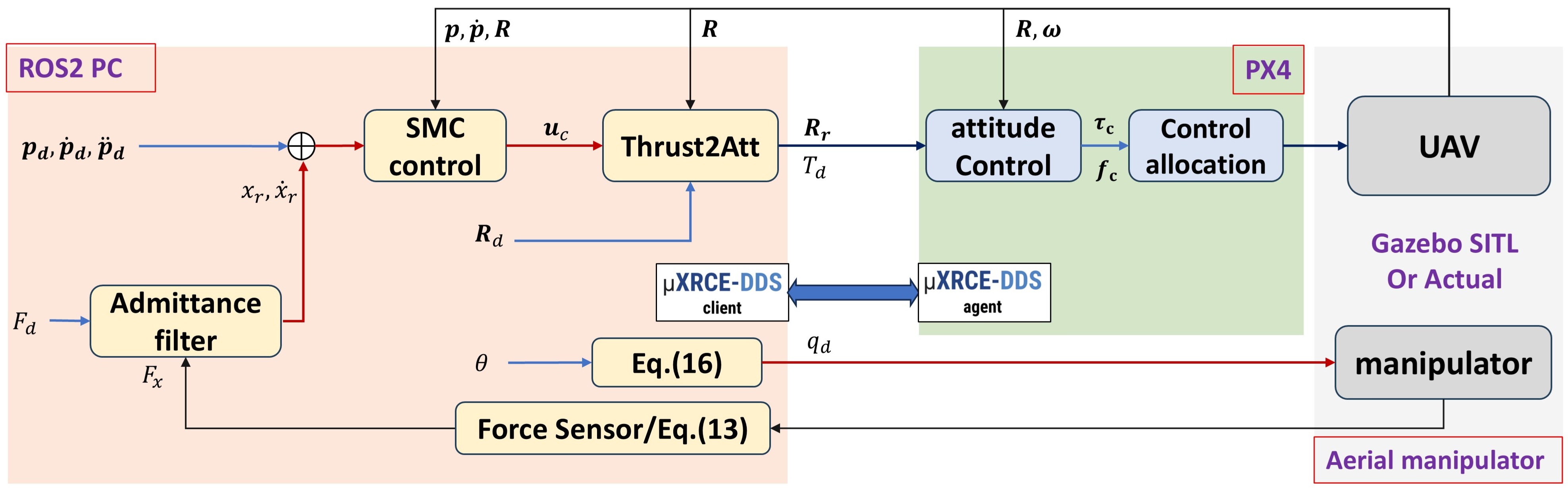}
\caption{Proposed system and control architecture}
\label{fig:sys_arch}
\end{figure*}

The system architecture (Fig. \ref{fig:sys_arch}) utilizes PX4 for geometric attitude control \cite{brescianiniNonlinearQuadrocopterAttitude2013} and control allocation. Motion, interaction, and manipulator control run on a ROS2 companion computer, sending orientation setpoints $\mathbf{R}_d$ and thrust commands $T_d$ via DDS middleware as proposed in \cite{alqutami2024modelingICOM}.

\section{Motion Control Design}
\subsection{Sliding mode control law}
SMC is a nonlinear robust control framework designed to ensure stable tracking in the presence of model uncertainties and matched disturbances. Its key principle is the enforcement of system trajectories onto a predefined sliding surface (manifold), on which the closed-loop dynamics exhibit reduced order and prescribed behavior that is insensitive to uncertainties. 

From the dynamics model \eqref{eq:newton_eom}, Define the virtual control vector $\mathbf{u}_c = \frac{\mathbf{R}\mathbf{f}_c}{m}$. The contact force is omitted and considered later in the interaction control law. The position and velocity tracking errors are $\mathbf{e}_p = \mathbf{p} - \mathbf{p}_d$ and $\dot{\mathbf{e}}_p = \dot{\mathbf{p}} - \dot{\mathbf{p}}_d$. The integral-augmented sliding surface $\mathbf{s}(t)$ and its derivative are:
\begin{IEEEeqnarray}{ll}
\mathbf{s} &= \dot{\mathbf{e}}_p + \boldsymbol{\Lambda} \mathbf{e}_p + \boldsymbol\eta \int_{0}^{t} \mathbf{e}_p(\tau) d\tau \label{eq:sliding_surface}\\
\dot{\mathbf{s}} &= \ddot{\mathbf{e}}_p + \boldsymbol{\Lambda}\dot{\mathbf{e}}_p + \boldsymbol\eta \mathbf{e}_p \label{eq:sliding_surface_derv}
\end{IEEEeqnarray}
where $\boldsymbol{\Lambda}, \boldsymbol{\eta} > 0$ are positive definite diagonal gain matrices.  $\int_{0}^{t} \mathbf{e}_p(\tau) d\tau$ is the additional integral term with integral gain $\boldsymbol\eta$.  \(\mathbf{\Lambda}\) governs the error convergence bandwidth while $\boldsymbol\eta$ eliminates  the steady-state errors under constant disturbances.

The control output $\mathbf{u}_c = \mathbf{u}_{eq} + \mathbf{u}_{sw}$ combines nominal equivalent control $\mathbf{u}_{eq}$ to handle nominal dynamics and switching control $\mathbf{u}_{sw}$ to provide robustness. Setting $\dot{\mathbf{s}}=\mathbf{0}$ yields:
\begin{equation}
   \begin{split}
       \dot{\mathbf{s}} = \ddot{\mathbf{e}_p} + \mathbf{\Lambda}\dot{\mathbf{e}_p} + \boldsymbol\eta \mathbf{e}_p &= 0 \\
       \left( \ddot{\mathbf{p}} - {\ddot{\mathbf{p}}}_{d} \right) + \mathbf{\Lambda}\dot{\mathbf{e_p}} + \boldsymbol\eta \mathbf{e}_p &= 0 \\
       \left( \mathbf{g}+ \mathbf{u}_{c} - {\ddot{\mathbf{p}}}_{d} \right) + \mathbf{\Lambda}\dot{\mathbf{e}} + \boldsymbol\eta \mathbf{e}_p  &= 0
   \end{split}
\end{equation}
Solving for \(\mathbf{u}_{c}\) to obtain the equivalent control \(\mathbf{u}_{eq}\) 
\begin{equation}
\mathbf{u}_{eq} =
\ddot{\mathbf{p}}_d - \boldsymbol{\Lambda} \dot{\mathbf{e}}_p - \boldsymbol\eta \mathbf{e}_p - \mathbf{g} 
\label{eq:u_eq}
\end{equation}

To dominate the disturbance bound $\sigma_f$, the switching term is defined with a positive definite diagonal gain matrix  $\mathbf{K} > \mathbf{0} \in \ \mathbb{R}^{3 \times 3\ }$:
\begin{equation}
\mathbf{u}_{sw} = - \mathbf{K} \text{sgn}(\mathbf{s}) \label{eq:smc_u_sw}
\end{equation}
where, \(sgn\left( \mathbf{s} \right)\) is the signum function acting element-wise.  This results in the complete virtual control law:
\begin{equation}
\mathbf{u}_c = \ddot{\mathbf{p}}_d - \boldsymbol{\Lambda} \dot{\mathbf{e}}_p - \boldsymbol\eta \mathbf{e}_p - \mathbf{g} - \mathbf{K} \text{sgn}(\mathbf{s}) \label{eq:u_smc}
\end{equation}

\subsection{SMC stability proof}
\textit{Theorem 1:} Control law \eqref{eq:u_smc} drives tracking errors to manifold $\mathbf{s}=\mathbf{0}$, ensuring asymptotic stability ($\lim_{t\to\infty} \mathbf{e}_p(t) = \mathbf{0}$).

\textit{Proof:} Consider candidate Lyapunov function $V(\mathbf{s}) = \frac{1}{2}\mathbf{s}^T\mathbf{s}$. Its time derivative along system trajectories, Substituting \eqref{eq:u_smc} and \eqref{eq:newton_eom}, gives:
\begin{equation}
\begin{aligned}
\dot{V} &= \mathbf{s}^T \left( \mathbf{g} + \mathbf{u}_c + \boldsymbol\Delta_f - \ddot{\mathbf{p}}_d + \boldsymbol\Lambda\dot{\mathbf{e}}_p + \boldsymbol\eta \mathbf{e}_p \right) \\
&= \mathbf{s}^T \left( -\mathbf{K}\text{sgn}(\mathbf{s}) + \boldsymbol\Delta_f \right) \\
&= - \mathbf{K} \|\mathbf{s}\| + \mathbf{s}^T\boldsymbol\Delta_f
\end{aligned}
\end{equation}
Using the Cauchy–Schwarz inequality \(\mathbf{s}^T\boldsymbol\Delta_f \leq \| \mathbf{s} \|\| \boldsymbol\Delta_f \| \leq \| \mathbf{s} \|\sigma_f\) and \(\mathbf{K}\| \mathbf{s} \| \geq 0\)
\begin{equation}    
\begin{aligned}
\dot{\mathbf{V}} & \leq \mathbf{- K}\| \mathbf{s} \| + \| \mathbf{s} \| \sigma_f \\
 & \leq \mathbf{-}\left( \mathbf{K -}\sigma_f \right)\| \mathbf{s} \|
\end{aligned}
\end{equation}
Selecting gain matrix $\mathbf{K}$ such that $\boldsymbol\delta = \mathbf{K} - \sigma_f > \mathbf{0}$ ensures:
\begin{equation}
\dot{V} \le -\boldsymbol\delta \|\mathbf{s}\|
\label{eq:Lyapunov_fcn_deriv}
\end{equation}
Eq. \eqref{eq:Lyapunov_fcn_deriv} satisfies the sliding condition $\dot{V} < 0$ for $\mathbf{s} \neq \mathbf{0}$, ensuring stability. On the manifold ($\mathbf{s}=\mathbf{0}$), linear differential dynamics $\ddot{\mathbf{e}}_p + \boldsymbol\Lambda\dot{\mathbf{e}}_p + \boldsymbol\eta \mathbf{e}_p = \mathbf{0}$ satisfy the Hurwitz stability criterion, proving $\mathbf{e}_p \to \mathbf{0}$ as $t \to \infty$.

\subsection{Chattering mitigation}
The discontinuous control action in \eqref{eq:smc_u_sw} can cause high-frequency chattering which may excite unmodeled dynamics and cause actuator saturation and wear, particularly during rigid surface contact and aggressive maneuvers. This is attenuated by replacing $\text{sgn}(\mathbf{s})$ in \eqref{eq:smc_u_sw} with a smooth hyperbolic tangent function, $\text{sgn}(\mathbf{s}) \approx \text{tanh}\left( \frac{\mathbf{s}}{\epsilon} \right)$, where $\epsilon$ defines the width of the boundary layer.

\section{Interaction Control}
In addition to motion control, regulating interaction forces is essential for safe and reliable inspection. A feedforward attitude augmented with an admittance controller is proposed for regulating the interaction force along the contact direction. In addition, the arm joint is used to compensate for the vehicle's tilt to exert force against the surface.

\subsection{Admittance-based force control}
Static force equilibrium along the contact direction relates force $F_x$, pitch angle $\theta$, and gravity:
\begin{equation}
F_x = -mg \tan(\theta) \implies \theta_d = -\arctan\left(\frac{F_d}{mg}\right) \label{eq:contact_force}
\end{equation}
The desired pitch angle $\theta_d$ required to achieve a target force $F_d$ is feedforward to the attitude controller. Furthermore, a physical force sensor directly captures contact interaction and dynamic coupling. 

To ensure smooth interaction, an admittance law shapes the dynamic relationship between motion and force and dynamically modifies trajectory based on force error \cite{alqutami2024modelingICEM}:
\begin{equation}
M_a \ddot{x}_r + D_a \dot{x}_r = F_d - F_x
\label{eq:admittance_law}
\end{equation}
where $M_a$ and $D_a$ denote virtual mass and damping and $F_x$ is the measured contact force. The stiffness term is set to zero since the position of the environment is assumed to be unknown; a constant forward velocity is used to approach the surface and establish contact. Reference trajectory $x_r, \dot{x}_r$ is updated via discrete integration:
\begin{equation}
\begin{aligned}
\ddot{x}_r[t] &= M_a^{-1}\left( F_d - F_x - D_a\dot{x}_r[t-1] \right) \\
\dot{x}_r[t] &= \dot{x}_r[t-1] + \ddot{x}_r[t]\Delta t \\
x_r[t] &= x_r[t-1] + \dot{x}_r[t]\Delta t
\end{aligned} \label{eq:admittance_dyn_proposed_acc}
\end{equation}
Admittance control triggers when measured contact force reaches $-1$~N and the reference motion is tracked by the inner SMC controller along the interaction force.

\subsection{Manipulator control}
To maintain sensor alignment against the surface, joint angle setpoint $q_d$ offsets vehicle pitch angle $\theta$:
\begin{equation}
q_d = \begin{cases} 
    -\theta & \text{if } |F_x| > 1~N \\
    0 & \text{otherwise}
\end{cases}
\label{eq:arm_cmd}
\end{equation}
This joint trajectory allows the arm to counteract UAV tilt by adjusting the contact geometry and improves contact stability during inspection. Joint motion is bounded by mechanical limits ($-\pi/2 \le q \le \pi/2$), while admittance filter smooths pitch transitions during contact onset. A rate limiter is also recommended to avoid large overshoots.

\subsection{Attitude projection}
The virtual command $\mathbf{u}_c$ is mapped to desired orientation matrix $\mathbf{R}_r = [\mathbf{x}_b, \mathbf{y}_b, \mathbf{z}_b] \in SO(3)$ and thrust $T_d$. Since the vehicle's thrust is always produced along its vertical body axis, the desired body z-axis is the normalized control vector
\begin{equation}
\mathbf{z}_b = \frac{\mathbf{u}_c}{\|\mathbf{u}_c\|}
\end{equation}

Projecting the desired yaw $\psi_d$ onto the horizontal plane to get the heading vector in $\mathcal{F}_I$, $\mathbf{x}_\psi = [\cos\psi_d, \sin\psi_d, 0]^T$ then constructing the orthonormal vector $\mathbf{y}_b$ using the cross product of heading vector and thrust direction
\begin{equation}
\mathbf{y}_b = \frac{\mathbf{z}_b \times \mathbf{x}_{\psi}}{\|\mathbf{z}_b \times \mathbf{x}_{\psi}\|}
\end{equation}
Finally, $\mathbf{x}_b$ is computed to complete the right-handed coordinate system
\begin{equation}
\mathbf{x}_b = \mathbf{y}_b \times \mathbf{z}_b
\end{equation}
The resulting rotation matrix $\mathbf{R}_{r}$ represents the transformation from $\mathcal{F}_I$ to $\mathcal{F}_B$  that is executed by the attitude controller.

\section{Results and Discussion}
System performance was evaluated in PX4-Gazebo SITL (ROS2, 80~Hz control loop) using parameters listed in Table \ref{tab:parameters}. A 5\% mass estimation error was injected to test controller robustness. Contact surface parameters were set to $10^7$~N/m stiffness and $0.1$~N$\cdot$s/m damping. Arm joint was controlled by PID to track the desired joint angle $q_d$.
\begin{table}[htbp]
\centering
\caption{Aerial Manipulator Parameters}
\label{tab:parameters}
\begin{tabular}{lcc}
\hline
\textbf{Parameter} & \textbf{Value} & \textbf{Unit} \\
\hline
UAV mass $m$ & 3.0 & kg \\
UAV inertia $\mathbf{J}$ & diag(0.08, 0.08, 0.12) & kg$\cdot$m$^2$ \\
UAV wheelbase & 0.48 & m \\
Arm length $r_e$ & 0.4 & m \\
Arm weight & 0.3 & kg \\
\hline
\end{tabular}
\end{table}
Control gains: $\mathbf{K} = \text{diag}(4.0, 5.0, 6.0)$, $\boldsymbol{\Lambda} = \text{diag}(5.5, 5.0, 4.0)$, $\boldsymbol{\eta} = \text{diag}(0.01, 0.01, 2.5)$, $\epsilon = 2.3$, $M_a = 5.0$~kg, $D_a = 0.8$~N$\cdot$s/m, obtained through empirical trial and error. 

Two scenarios have been used to evaluate the performance of the proposed system: free-flight and contact inspection scenarios. Performance metrics used are Root Mean Squared Errors (RMSE), error norm $\|\mathbf{e}_p\|=\sqrt{e_x^2+e_y^2+e_z^2}$, and control input norm $\|\mathbf{u}_c\|$. 

\subsection{Free flight tracking and coupling rejection}
Free-flight motion tracking was tested along trajectory $\mathbf{p}_d(t) = [2.0\sin(\frac{\pi}{8}t), 1.0\sin(\frac{\pi}{4}t), 1.0\sin(\frac{\pi}{8}t)+2.0]^T$~m with simultaneous arm motion $q_d(t) = 0.5\sin(\frac{\pi}{6}t)$~rad. The desired yaw $\psi_d$ is kept zero.

To demonstrate the performance of the proposed SMC controller, it is compared with classical PD control with inverse-dynamics compensation. The control parameters are set to $K_p=[7.5, 7.5, 13.0]$, and $K_d=[5.0, 5.0, 7.0]$.
\begin{figure*}[htbp]
\centering
\includegraphics[width=1\textwidth]{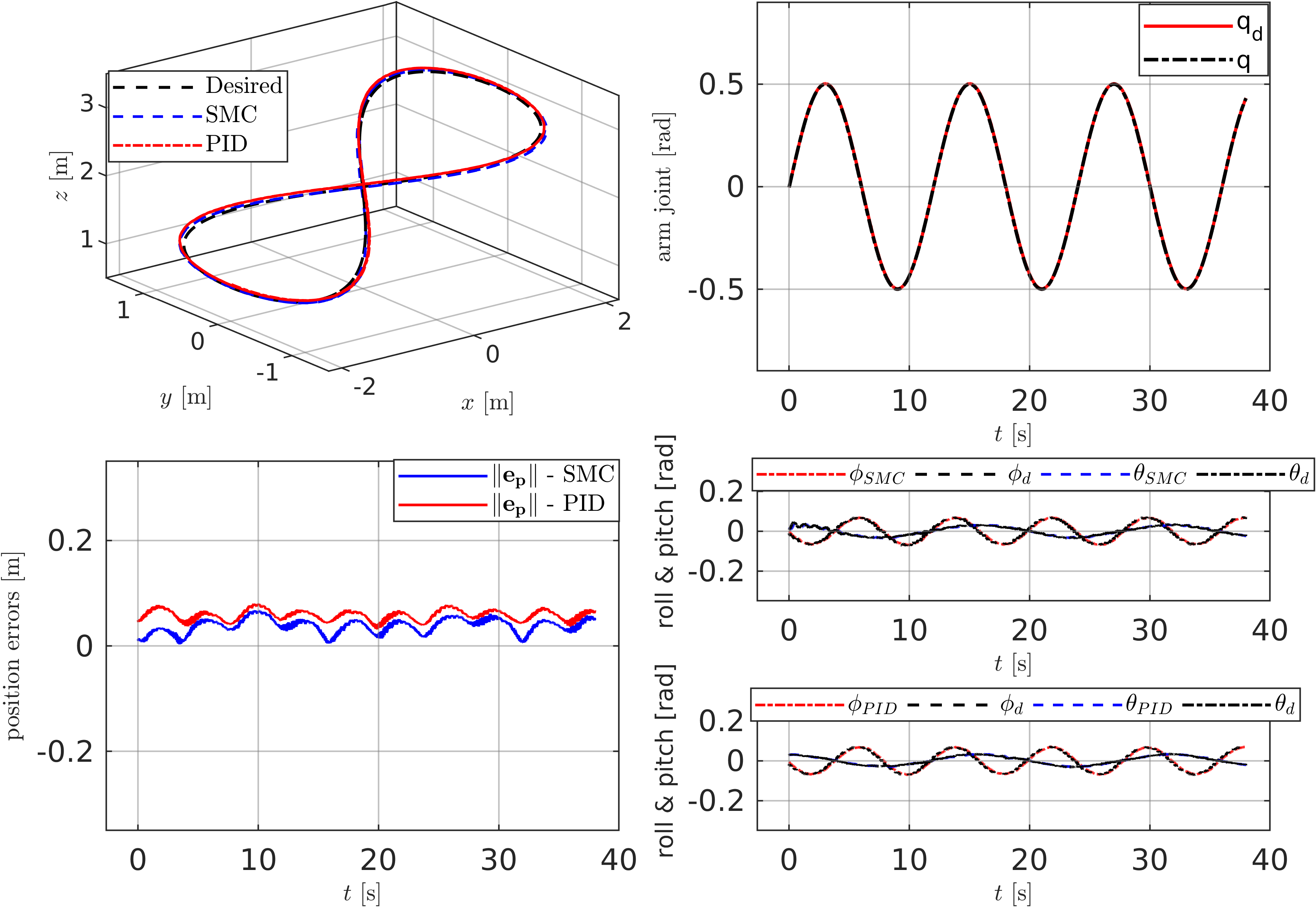}
\caption{Position tracking for SMC and PID.}
\label{fig:fig8_response}
\end{figure*}

As shown in Fig. \ref{fig:fig8_response} and the performance summary in Table \ref{tab:performance_summary}, SMC achieves lower tracking errors than  PID control. Despite periodic arm motion (top-right), the controllers maintained stable trajectory tracking (top-left).
\begin{table}[htbp]
\caption{Performance summary for trajectory tracking}
\begin{center}
\begin{tabular}{cccccc}
\hline
\textbf{Controller} & x-RMSE & y-RMSE & z-RMSE & $\|\mathbf{e_R}\|$ & $\|\mathbf{u}_c\|$\\
\hline
\textbf{SMC} & 0.021 & 0.034 & 0.004 & 0.01 &  0.698  \\
\textbf{PID} & 0.023 & 0.039 & 0.038 & 0.01 & 0.693 \\
\end{tabular}
\label{tab:performance_summary}
\end{center}
\end{table}

\begin{figure*}[htbp]
\centering
\includegraphics[width=0.95\textwidth]{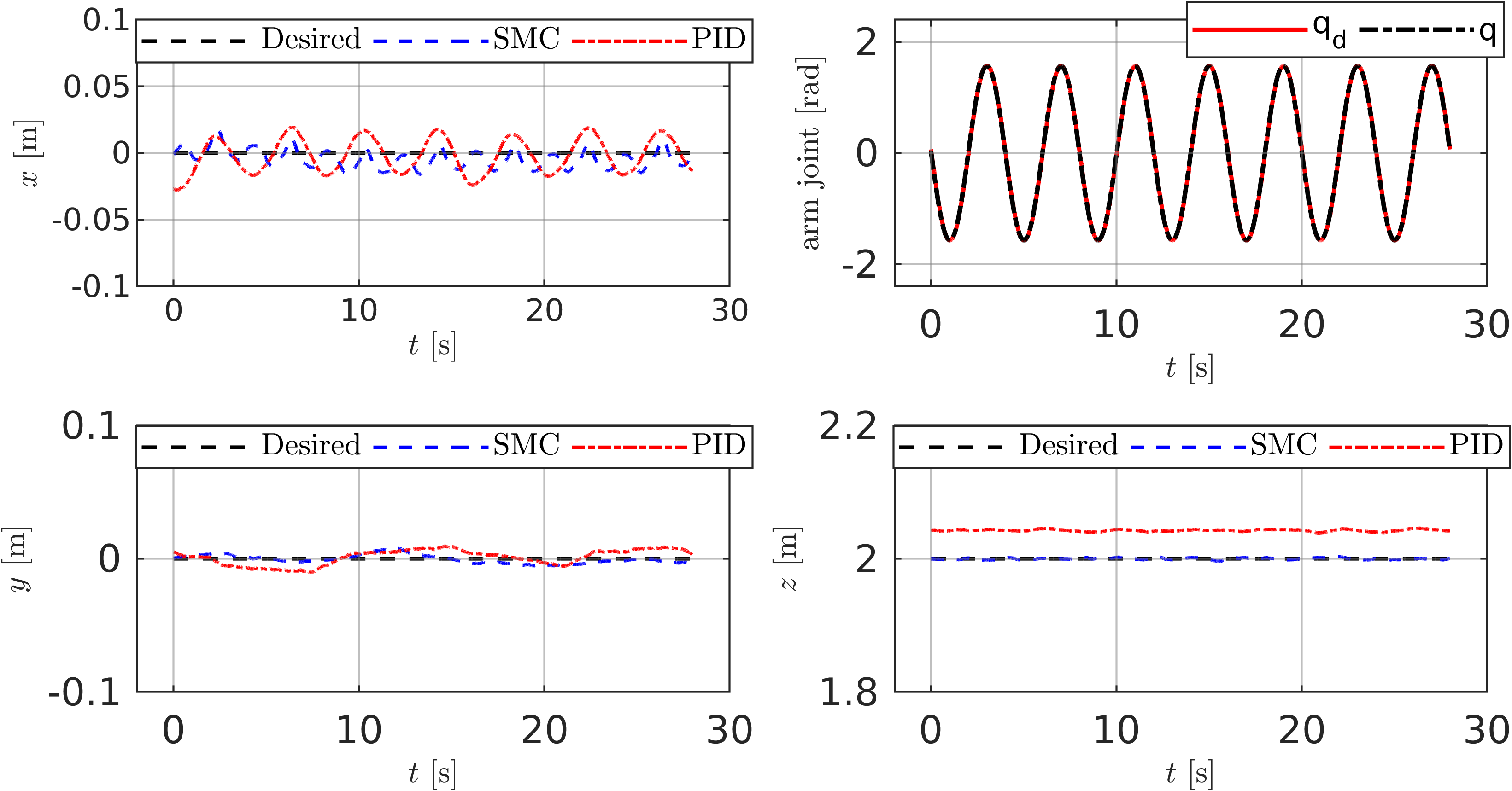}
\caption{Position hold performance of PID and SMC under arm swing disturbance}
\label{fig:pos_hold}
\end{figure*}

Robustness was further tested by forcing the UAV to hover while subject to large arm swings $q_d(t) = \frac{\pi}{2}\sin(\frac{\pi}{2}t)$~rad (Fig. \ref{fig:pos_hold}). SMC suppressed dynamic coupling significantly better than PID. The RMSE of the x-position, the most affected by arm motion, is almost double for PID compared to SMC. The position error norm for PID is 4.5~cm, while the error norm for SMC is only 0.8~cm.

\subsection{Contact interaction}
\begin{figure}[!ht]
\centerline{\includegraphics[width=1\linewidth]{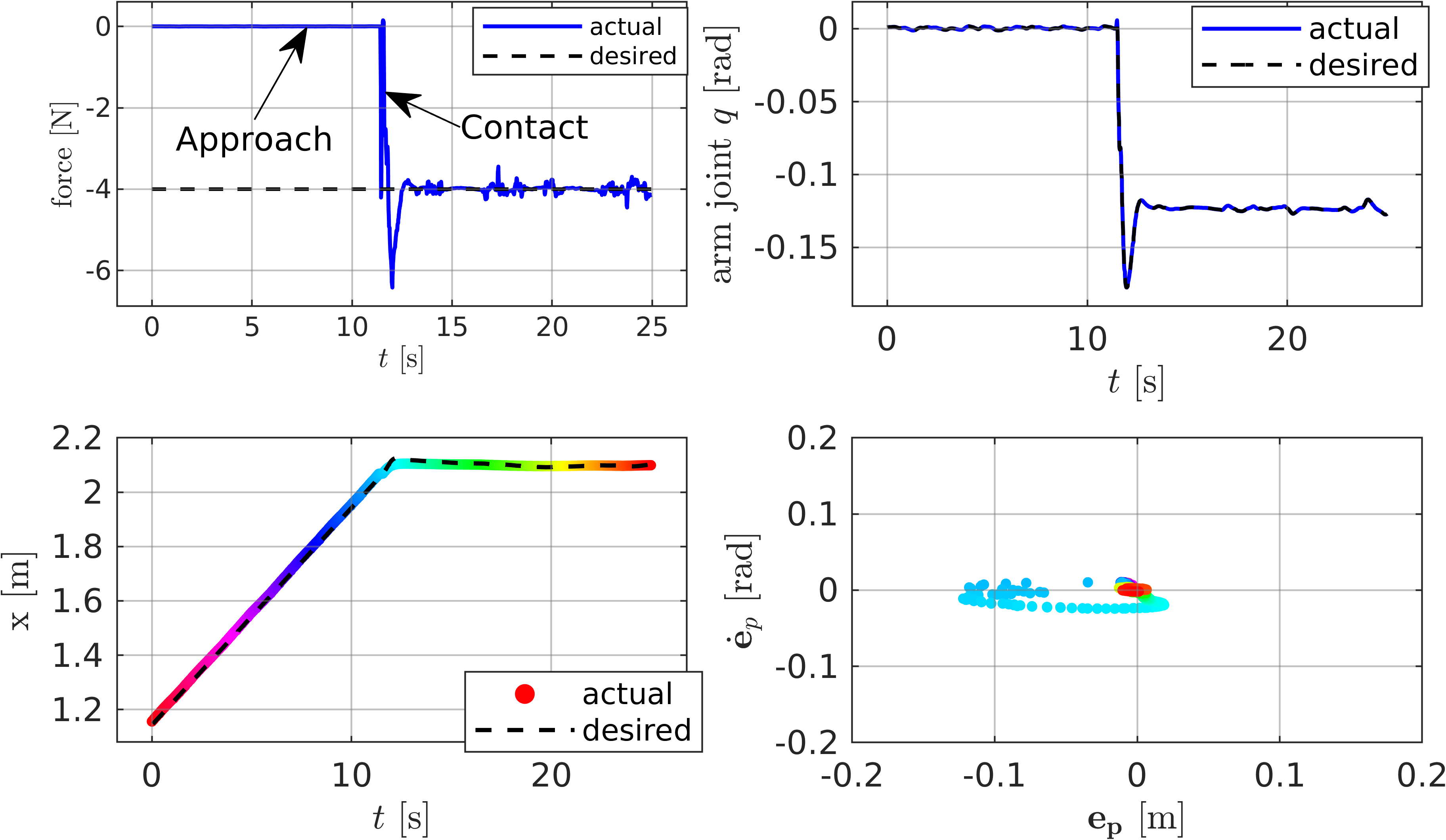}}
\caption{Approach and establish contact}
\label{fig:contact}
\end{figure}

The second scenario evaluates the proposed pitch-compensation and interaction control using the proposed aerial manipulator. The vehicle approaches the surface at constant velocity of 8~cm/s, establishes contact, and exerts a setpoint force $F_d = -4$~N as shown in Fig.~\ref{fig:contact}.

This scenario demonstrates the effectiveness of the proposed mechanism in maintaining stable contact. The UAV maintained a pitch angle of approximately $7^\circ$ to achieve the -4~N pushing force, which was compensated by joint $q_d$ to preserve surface alignment. The x-position response, bottom-left plot, is color-coded with time to show how the SMC phase-plane (error vs error derivative in the bottom right plot) maintained the sliding state at the origin and quickly pulled the sliding surface back to the origin when the position setpoint changed at contact. The SMC maintained a UAV error norm of 1~cm for the yz-axes.

\begin{figure}[!ht]
\centerline{\includegraphics[width=1\linewidth]{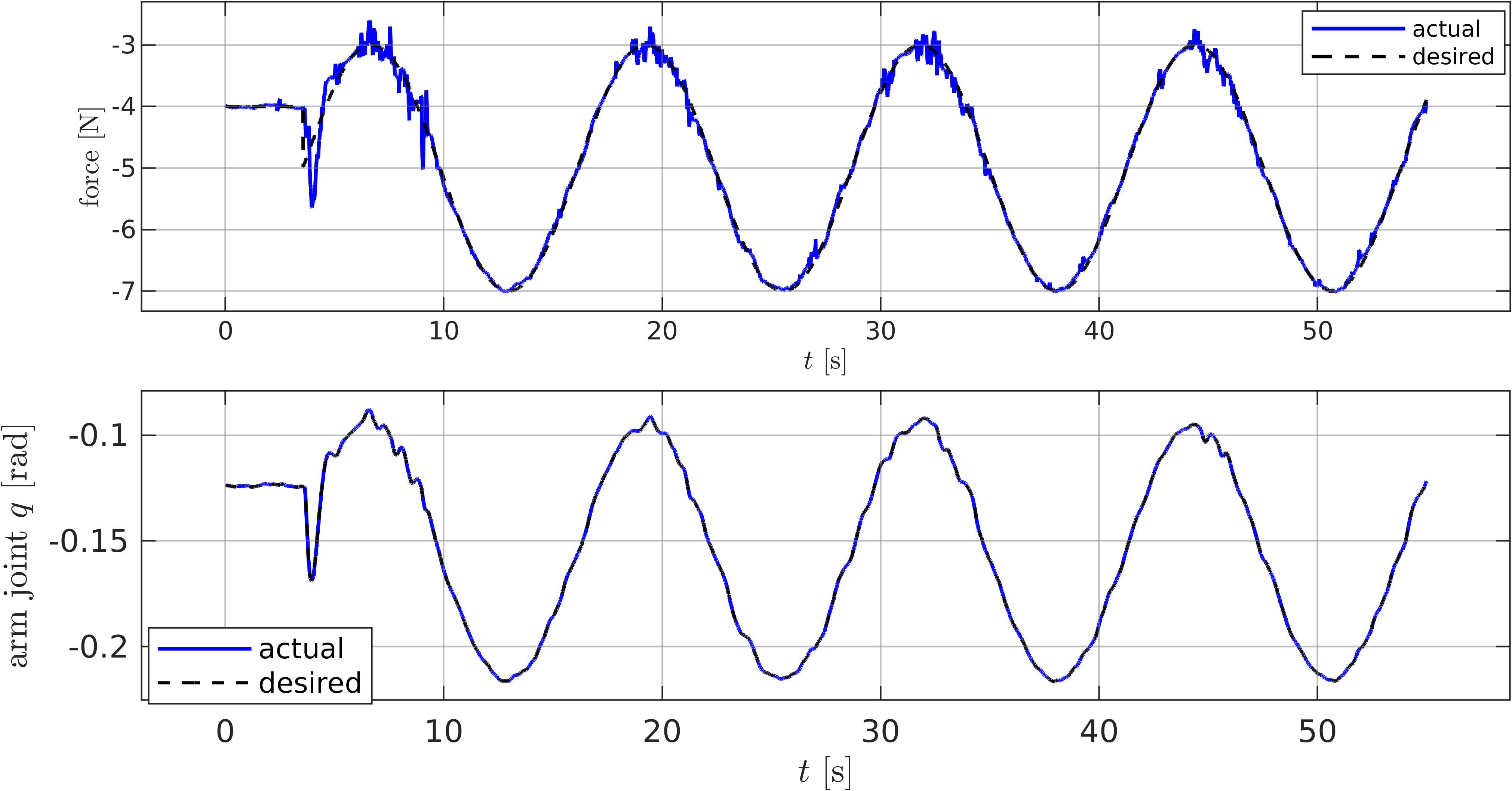}}
\caption{Contact force tracking performance.}
\label{fig:force_track}
\end{figure}

Dynamic force tracking was evaluated using sinusoidal command $F_d(t) = 2\sin(\frac{\pi}{6}t) - 5$~N. The force tracking response and corresponding arm joint compensation are shown in Fig.~\ref{fig:force_track}. The controller tracked the target force accurately (RMSE = 0.12~N) with minimal position drift ($y\text{-RMSE} = 0.009\text{~m}, z\text{-RMSE} = 0.003\text{~m}$) and smooth control effort ($\|\mathbf{u}_c\| = 2.46$).

\begin{figure}[!ht]
\centerline{\includegraphics[width=1\linewidth]{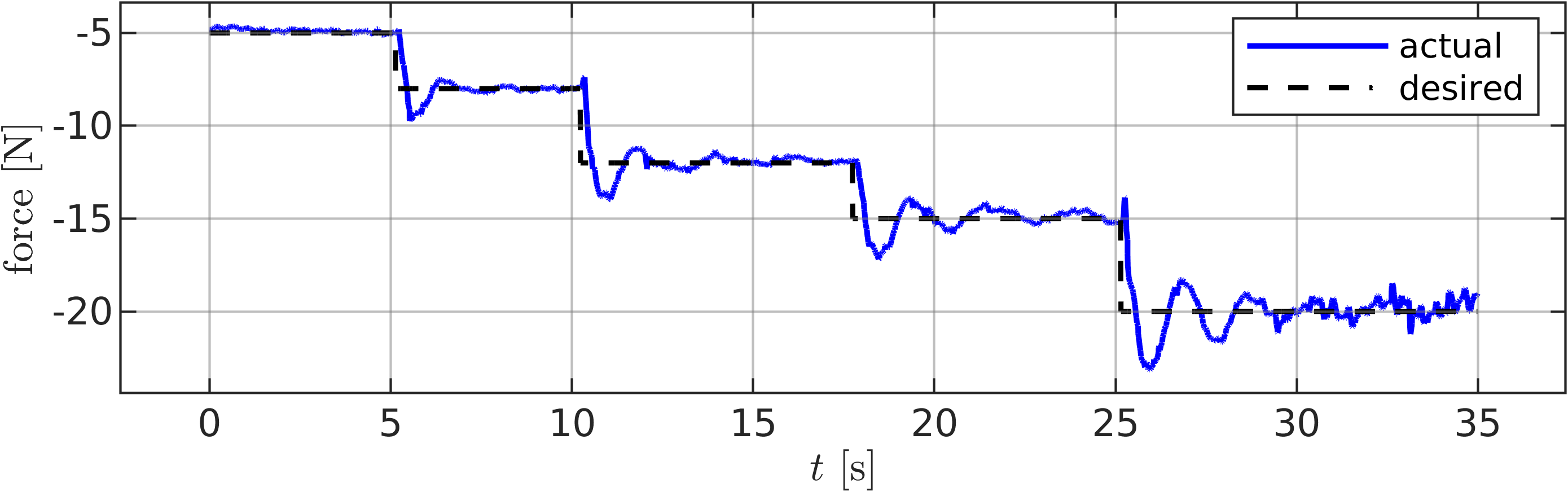}}
\caption{Contact force step response with active pitch compensation.}
\label{fig:step_force_track}
\end{figure}

Finally, to evaluate the system limit, the force setpoint was increased in steps until attitude stability was lost. The controller was able to exert up to 20~N of contact force, shown in Fig.~\ref{fig:step_force_track}, with a corresponding pitch angle of 31~degrees and a 0.88~N force RMSE.  Without active arm compensation ($q_d = 0$), the vehicle lost stability past 9~N, proving the importance of active joint alignment.

\FloatBarrier
\section{Conclusion and Future Work}
This paper presented a robust control architecture for an underactuated aerial manipulator. By combining SMC with admittance-based force regulation, the system effectively manages the dynamic coupling between the UAV and its 1-DoF manipulator. Results confirmed that the proposed scheme outperforms classical PID control in tracking accuracy and disturbance rejection, achieving a force regulation RMSE of 0.12 N while maintaining surface alignment through active pitch compensation.

Future work will focus on experimental validation on a physical prototype and extend the design to support push-and-slide applications, such as cleaning, and to handle different surface geometries.

\bibliographystyle{IEEEtran}
\bibliography{refs}    % Points to references.bib

\end{document}